\documentclass[11pt,a4paper]{article}
\usepackage[dvips]{graphicx}
\usepackage{subfigure}
\usepackage{amsmath}

\newcommand{\tenss}[1]{\mbox{\boldmath$#1$}}
\newcommand{\vek}[1]{\mathbf{#1}}

\newcommand{\stress}{\sigma}
\newcommand{\strain}{\varepsilon}

\title{Back Analysis of Microplane Model Parameters Using Soft
Computing Methods}
\author{%
Anna Ku\v{c}erov\'a
\thanks{%
Corresponding author,
Address:~Department of Mechanics, 
Faculty of Civil Engineering,
Czech Technical University in Prague,
Th\'{a}kurova 7, 166 29 Prague 6, Czech Republic
}
\and 
Mat\v{e}j Lep\v{s}
\and 
Jan Zeman
}
\date{}

\begin{document}

\maketitle

\begin{abstract}
A~new procedure based on layered feed-forward neural networks for the
microplane material model parameters identification is proposed in the
present paper. Novelties are usage of the Latin Hypercube Sampling
method for the generation of training sets, a~ systematic employment
of stochastic sensitivity analysis and a~genetic algorithm-based
training of a~neural network by an~evolutionary algorithm. Advantages
and disadvantages of this approach together with possible extensions
are thoroughly discussed and analyzed.
\end{abstract}

\section{Introduction}
%
Concrete is one of the most frequently used materials in Civil
Engineering. Nevertheless, as a~highly heterogeneous material, it
shows very complex non-linear behavior, which is extremely difficult
to describe by a~sound constitutive law. As a~consequence, numerical
simulation of response of complex concrete structures still remains a
very challenging and demanding topic in engineering computational
modeling.  

One of the most promising approaches to modeling of concrete behavior
is based on the microplane concept, see,
e.g.,~\cite[Chapter~25]{Jirasek:2001:IAS} for general exposition
and~\cite{Bazant:2005:M5} for the most recent version of this family
of models. It leads a~fully three-dimensional material law that
incorporates tensional and compressive softening, damage of the
material, supports different combinations of loading, unloading and
cyclic loading along with the development of damage-induced anisotropy
of the material. As a~result, the M4 variant of the microplane model
introduced in~\cite{Bazant:2000:M4} is fully capable of predicting
behavior of real-world concrete structures once provided with proper
input data, see~\cite{Nemecek:2004:EI,Nemecek:2005:ET} for concrete
engineering examples. The major disadvantages of this model are,
however, a~large number of phenomenological material parameters and
a~high computational cost associated with structural analysis even in
a parallel implementation~\cite{Nemecek:2002:MM}. Although the authors
of the model proposed a~heuristic calibration procedure~\cite[Part
II]{Bazant:2000:M4}, it is based on the trial-and-error method and
provides only a~rude guide for determination of selected material
parameters.  Therefore, a~reliable and inexpensive procedure for the
identification of these parameters is on demand.

In the view of potential improvements demonstrated in the recent work
by Nov\'{a}k and Lehk\'{y}~\cite{Novak:2004:NNB}, the applicability of
a~novel procedure based on artificial neural networks~(ANN's) for the
microplane parameter identification is examined in the present
contribution. Individual steps of the identification procedure
involve~(see also \cite{Novak:2004:NNB,Strauss:2004:SP}
for more details)
\begin{description}

\item[Step~1] {\em Setup of} a~virtual and/or real experimental {\em test}
  used for the identification procedure.

\item[Step~2] Formulation of an appropriate computational model. {\em Input
  data} to the model coincide with the parameters to be identified.

\item[Step~3] {\em Randomization} of input parameters. Input data are
  typically assumed to be random variables uniformly distributed on a
  given interval. 

\item[Step~4] Stochastic {\em sensitivity analysis} using the Monte
  Carlo-based simulation. This provides us with {\em relevant model
  parameters} which can be reliably identified from the computational
  simulation.

\item[Step~5] Definition of {\em topology} of an ANN used for the
  identification procedure.

\item[Step~6] {\em Training} of the ANN. The training set is formed from the
  data generated during the sensitivity analysis step.

\item[Step~7] {\em Validation} of the ANN with respect to the
  computational model. This step is usually performed by comparing the
  prediction of the ANN with an independent set of input
  data\footnote{usually called a ``training set''}.

\item[Step~8] The identification of relevant model parameters using
  trained ANN from available {\em experimental data}.

\end{description}

\noindent
In the rest of the paper a~more detailed description of the individual
steps when applied to the microplane~M4 model is presented. The basic
outline of the constitutive model together with numerical solution
scheme are presented in Section~2. The randomization of input
parameters using small-sample Monte Carlo simulation is described in
Section~3 together with stochastic sensitivity analysis. Section~4 is
devoted to the procedure of ANN's training. The application to
the identification of microplane model parameters is introduced in
Section~5. Finally, the results obtained using the methodology are
summarized in Section~6 together with comments on possible improvements.

\section{Microplane model M4 for concrete}
\begin{figure}[th]
\centering
\includegraphics[width=5cm,keepaspectratio]{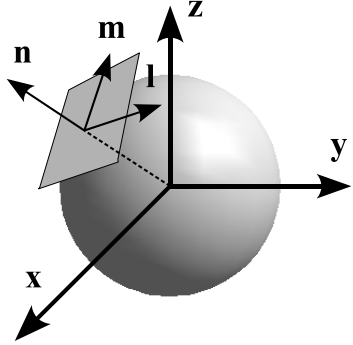}
\caption{Concept of microplane modeling}
\label{fig:microplane}
\end{figure}

%
In contrary to traditional approaches to constitutive modeling, which
build on description via second-order strain and stress {\em tensors}
at individual points in the $( x, y, z )$ coordinate system, the
microplane approach builds the descriptions on planes of arbitrary
spatial orientations -- so-called {\em microplanes}, related to a
macroscopic point, see Figure~\ref{fig:microplane}. This allows to
formulate constitutive equations in terms of stress and strain {\em
vectors} in the coordinate system $( \vek{l}, \vek{m}, \vek{n} )$
associated with a~microplane oriented by a~normal vector
$\vek{n}$. The general procedure of evaluation of a~strain-driven
microplane model response for a~given ``macroscopic'' strain tensor
$\tenss{\strain}( \vek{x} )$ can be described as follows: (i)~for a
given microplane orientation $\vek{n}$ normal ``macroscopic'' strain
tensor $\tenss{\strain}( \vek{x} )$ is projected onto the normal
``microstrain'' vector $\vek{\strain}( \vek{n})$ and the shear
microstrains $\vek{\strain}( \vek{m} )$ and $\vek{\strain}( \vek{l}
)$, (ii)~the normal and shear microstresses $\vek{\stress}( \vek{n} ),
\vek{\stress}( \vek{m} )$ and $\vek{\stress}( \vek{l} )$ are evaluated
using microplane constitutive relations, (iii)~the ``macroscopic''
stress tensor $\tenss{\stress}( \vek{x} )$ is reconstructed from the
microscopic ones using the principle of virtual work, see,
e.g.,~\cite[Chapter~25]{Jirasek:2001:IAS} for more details. In the
particular implementation, 28 microplanes with a~pre-defined
orientation on the unit hemisphere is used to evaluate the response of
the model.

To close the microplane model description, the appropriate microplane
constitutive relation must be provided to realistically describe
material behavior. The model examined in the current work is the
microplane model M4~\cite{Bazant:2000:M4}. The model uses
volumetric-deviatoric split of the normal components of the stress
and strain vectors, treats independently shear components of a
microplane and introduces the concept of ``boundary curves'' to limit
unrealistically high values predicted by earlier version of the
model. As a~result, the strain-to-stress map $\tenss{\strain}( \vek{x}
) \mapsto \tenss{\stress}( \vek{x} )$ is no longer smooth, which
complicates the formulation of consistent tangent stiffness
matrix~\cite{Nemecek:2002:MM} and, subsequently, gradient-based
approaches to material model parameters identification.

In overall, the microplane model~M4 needs eight parameters to describe
a~certain type of~concrete, namely: Young's modulus $E$, Poisson's
ratio $\nu$, and other six parameters ($k_1$, $k_2$, $k_3$, $k_4$,
$c_3$, $c_{20}$), which do not have a~simple physical interpretation,
and therefore it is difficult to determine their values from
experiments. The only information available in the open literature are
the bounds shown in the Table~\ref{tbl1}.

\begin{table}[ht]
\centering
\begin{tabular}{rcl}
\hline
{\bf Parameter} & & {\bf Bounds} \\
\hline
$E$      & $\in$ & $\langle 20.0, 50.0 \rangle$ \mbox{ GPa} \\
$\nu$    & $\in$ & $\langle 0.1, 0.3 \rangle$   \\
$k_1$    & $\in$ & $\langle 0.00008, 0.00025 \rangle$   \\
$k_2$    & $\in$ & $\langle 100.0, 1000.0 \rangle$ \\
$k_3$    & $\in$ & $\langle 5.0, 15.0 \rangle$ \\
$k_4$    & $\in$ & $\langle 30.0, 200.0 \rangle$ \\
$c_3$    & $\in$ & $\langle 3.0, 5.0 \rangle$ \\
$c_{20}$ & $\in$ & $\langle 0.2, 5.0 \rangle$ \\
\hline
\end{tabular}
\caption{Bounds for the microplane model parameters}
\label{tbl1}
\end{table}

In the present work, the computational model of a~structure is
provided by the object-oriented C++ finite element code {\bf
OOFEM~1.5}~\cite{Patzak:2001:OOFEM}. Spatial discretization is
performed using linear brick elements with eight integration
points. The arc-length method with elastic stiffness matrix is used to
determine the load-displacement curve related to the analyzed
experiment.\\

\section{Input parameter randomization and stochastic sensitivity
analysis }
%
The novelty of the identification approach proposed
in~\cite{Novak:2004:NNB} is a~systematic use of small-sample Monte
Carlo simulation method for generation of neural network training sets
as well as stochastic sensitivity analysis. In the particular case of
the M4 microplane model, each input parameter is assumed to be
uniformly distributed on an interval specified in Table~\ref{tbl1}.

The Latin Hypercube Sampling~(LHS) method~\cite{Iman:1980:SSS} is used
to generate particular realization of input variables as it enables to
minimize the amount of simulations needed to reliably train a~neural
network. Moreover, the Simulated Annealing optimization method
available in the software package FREET~\cite{Novak:2002:FREET} is
used to maximize the statistical independence among individual
samples. The Pearson product moment correlation coefficient, defined
as
\begin{equation}
cor = \frac{\sum(x_i-\bar{x})(y_i-\bar{y})}
{\sqrt{\sum(x_i-\bar{x})^2\sum(y_i-\bar{y})^2}},
\end{equation}
where $\bar{x}$ and $\bar{y}$ denote the expected values of random
variables $X$ and $Y$, is used as a~sensitivity measure to
investigate the influence of individual parameters to a~structural
response. Note that the correlation coefficient is normalized as $ -1
\leq cor \leq 1$, where higher absolute values indicate statistical
dependence of the random output variable $Y$ on the random input
variable $X$. \\

\section{Artificial neural network}
In this work, layered fully connected feed-forward neural networks
with~bias neurons~(see, e.g,~\cite{Tsoukalas:1997:FNA}) are used for
the parameter identification.
In general, a~neural network is~created to~map the input vector $I =
(I_0, I_1, \dots I_m)$ on a~target vector $T = (T_0, T_1, \dots T_n)$.
There are $L$ layers denoted as~$l_0, l_1 \dots l_{L-1}$, where $l_0$
is the input layer and~$l_{L-1}$ is the output layer.  The $i$-th
layer~$l_i$ has $N_i$ neurons denoted as~$n_{i,1}, n_{i,2}, \dots
n_{i,N_i}$.  Each layer except the output layer has the bias
neuron~$n_{i,0}$.  The connections are described by the weights
$w_{l,i,j}$, where $l = 1, 2 \dots L-1$ denotes a~layer, $i = 0, 1
\dots N_{l-1}$ is the index number of a~neuron in~the preceding
layer~$l-1$ ($i = 0$ for~bias neurons) and~$j = 1, 2 \dots N_l$ is the
index number of~a~neuron in~the layer~$l$. The output of~the neuron
$n_{l,j}$ is then defined as
\begin{eqnarray}
O_{l, j} &=& f_{\rm act}\left(\sum_{i=0}^{N_{l-1}}
O_{l-1,i}\,.\,w_{l,i,j}\right), \;
l = 1,2\dots L-1, \; j = 1,2\dots N_l\, , \\
O_{0, j} &=& I_j,\; j = 1,2\dots N_0\, ,\\
O_{l, 0} &=& 1, \; l = 0,1\dots L-1\, ,
\end{eqnarray}
where $f_{\rm act}$ is an~activation function. In our current
implementation the activation function has the following form:
\begin{equation}
f_{\rm act}(\Sigma) = \frac{1}{( 1 + e^{ -\alpha/\Sigma } )}\, ,
\end{equation}
where $\alpha$ is the gain of~the $f_{\rm act}$. The value $\alpha =
0.5$ is used in all reported calculations. The output vector of each
layer $l_i$ is denoted as $O_{i} = (O_{i,1}, O_{i,2}, \dots O_{i,
  N_i})$.  Finally, the neural network is propagated as follows:
\begin {enumerate}
\item Let $l=1$.
\item Calculate $O_{l,i}$ for~$i = 1,2\dots N_l$. \label{prop1}
\item $l = l+1$.
\item If $l < L$ go to~\ref{prop1}, else $O_{L-1}$ is the network's
 approximation of~$T$.
\end{enumerate}
The output error, which is used as~a measure of a~training level, is
defined as
\begin{equation}
\label{Eq_Error}
	\varepsilon = \sqrt{\sum_{i=1}^{N_{L-1}} \left(T_i - O_{L-1, i}
 \right)^2}\, .
\end{equation}

\subsection{Training algorithm}\label{sec:GRADE}
%
Behavior of a~neural network is determined by a~preceding training
process. It consists of~finding the synaptic weights, which have
influence on~the response of a~neural network, depending on~the
different components of~an~input signal. The training of~a~neural
network itself could be considered as~an~optimization process, because
it can be seen as~a~minimization of~neural network output
error~(\ref{Eq_Error}). Then the synaptic weights of~a~neural network
act as~variables of~the optimization algorithm's fitness function. As
it was shown earlier, e.g. in~\cite{Drchal:2003:CTU}, evolutionary
algorithm-based optimizers can significantly outperform the
traditional methods, e.g. the backpropagation method. Therefore, the
evolutionary optimization algorithm {\bf GRADE}, proposed
in~\cite{Ibrahimbegovic:2003:IJNME}, is used for the neural network
training. Due to size limitations, we present only a~sketchy
description of the corresponding procedure and refer an interested
reader to~\cite{Ibrahimbegovic:2003:IJNME,Hrstka:2004:AES} for a~more
elaborate discussion.

In the computations to follow we will work with the population of $10
\times n$ chromosomes, where $n$ is the total number of unknowns in
the problem. This population evolves through the following operations:

\begin{description}

\item[Mutation] Let ${\bf x}_i(g)$ be the {\em i}-th chromosome in
a~generation~{\em g},
\begin{equation}
{\bf x}_i(g)=(x_{i1}(g),x_{i2}(g),...,x_{in}(g)),
\end{equation}
where $n$ is the number of variables of the objective function.  If
a~certain chromosome ${\bf x}_i(g)$ is chosen to be mutated, a~random
chromosome $RP$ is generated from the definition domain and a~new one
${\bf x}_k(g+1)$ is computed using the following relation:
\begin{equation}
{\bf x}_k(g+1)={\bf x}_i(g)+MR(RP-{\bf x}_i(g)).
\end{equation}
Parameter $MR$ is chosen randomly from the interval $(0,1)$. The
number of new chromosomes created by the mutation operator is defined
by 'radioactivity', which is a~parameter of the algorithm, with
a~constant value set to $0.2$ for all reported calculations.

\item[Gradient cross-over] The aim of~the cross-over operator is
to~create as many new chromosomes as there were in~the last
generation. The operator creates new chromosome ${\bf x}_i(g+1)$
according to the following sequential scheme: choose randomly two
chromosomes ${\bf x}_q(g)$ and ${\bf x}_r(g)$, compute their
difference vector, multiply it by~a~coefficient $CR$ and add it to~the
better one of them, i.e,
\begin{equation}
{\bf x}_i(g+1)=max({\bf x}_q(g);{\bf x}_r(g)) + SG \times CR({\bf
x}_q(g)-{\bf x}_r(g)).
\label{eq_cross_grade}
\end{equation}
The parameter $CR$ is chosen randomly from the interval $(0, CL)$,
where $CL$ is a~parameter of the algorithm equal to $1.0$ for all our
calculations. $SG$ denotes the sign change parameter which is supposed
to get the correct orientation of the increase with respect to the
gradient of the objective function.

\item[Selection] represents the kernel of each genetic algorithm. The
goal is to~provide a~progressive improvement of~the whole population,
which is achieved by~reducing the number of~the ``living'' chromosomes
together with conservation of the better ones. Modified tournament
strategy is used for~this purpose: two chromosomes are chosen randomly
from a~population, they are compared and~the worse of~them is cast
off. This conserves population diversity thanks to~a~good chance
of~survival even for~badly performing chromosomes.

\end{description}

Moreover, sometimes the method is ``caught'' in a local extreme and
has no chance to escape unless a mutation randomly finds a sub-area
with better values. If the gradient optimization methods are applied,
this case is usually resolved by so-called {\em multi-start}
principle. It consists of restarting the algorithm many times with
different starting points. Similarly, any type of an evolutionary
algorithm could be restarted many times. Nevertheless, the experience
shows that there are functions with so-called deceptive behavior (and
the training of a neural network is one of them), characterized by
a~high probability that the restarted algorithm would fall again into
the same local extreme rather than explore another sub-area.

As a solution, the CERAF\footnote{Abbreviation of the French
expression {\em CEntre RAdioactiF} - the radioactivity center.} method
has been introduced in~\cite{Hrstka:2004:AES}. It produces areas of
higher level of ``radioactivity'' in the neighborhood of all
previously found local extremes by substantially increasing the
mutation probability in these areas~(this probability is set to
$100\%$ hereafter). The diameter of the radioactivity area (finally it
defines a~{\em n}-dimensional hyper-ellipsoid for all variables) is
set to a~$75\%$ percentage of an appropriate variable interval. The
time of stagnation that precedes the markup of a local extreme is a
parameter of the method set to $100$ generations. Similarly to the
living nature, the radioactivity in the CERAF method is not constant
in time but decreases during the time as the solutions produced by the
cross-over are trying to get inside the radioactivity area, see
again~\cite{Hrstka:2004:AES} for more details. When the number of
stagnating generations is determined (the change between two best
solutions in following generations is less than some very small value,
$5\cdot10^{-11}$ in our case), the actual best solution is declared as
the center of the new radioactive area and the whole population is
restarted.

\section{Identification of microplane model M4 parameters}
%
The present section summarizes the individual steps of M4~material
model identification. Following the heuristic calibration procedure
suggested in~\cite[Part II]{Bazant:2000:M4}, we examine three specific
experimental tests: (i)~uniaxial compression, (ii)~hydrostatic test
and (iii)~triaxial test. Advantage of these tests is their simplicity
and availability in most experimental facilities. Moreover, authors
in~\cite{Bazant:2000:M4} claim that these experiments are sufficient
to determine all parameters of the microplane model M4. The results
presented in this section can be understood as a~verification of this
claim. \\


\begin{figure}[tb]
\centering
\includegraphics[height=5cm]{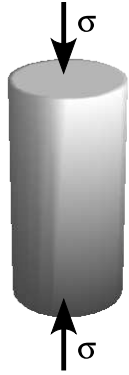}
\caption{Uniaxial test - Experiment setup} 
\label{fig:unixial_compression}
\end{figure}

\subsection{Uniaxial compression test}\label{sec:uni}
%
The most common experiment used for the determination of concrete
parameters is the uniaxial compression test on cylindrical concrete
specimens. In particular, the cylinder with a~radius equal to 75~mm
and the height of 300~mm is used. The set-up of the experiment is
shown in Figure~\ref{fig:unixial_compression}.

\vspace{6mm}
\begin{figure}[tb]
\centering
\includegraphics*[width=8cm,keepaspectratio]{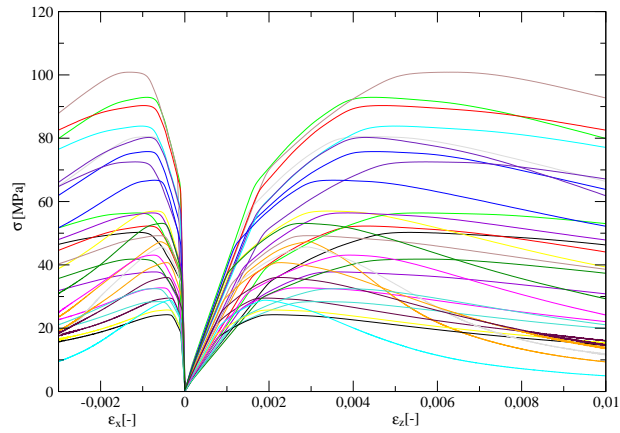}
\caption{Bundle of simulated stress-strain curves for uniaxial
compression test}
\label{fig:uniaxial_compression_lhs}
\end{figure}

\begin{figure}[tb]
\centering
\includegraphics*[width=14cm,keepaspectratio]{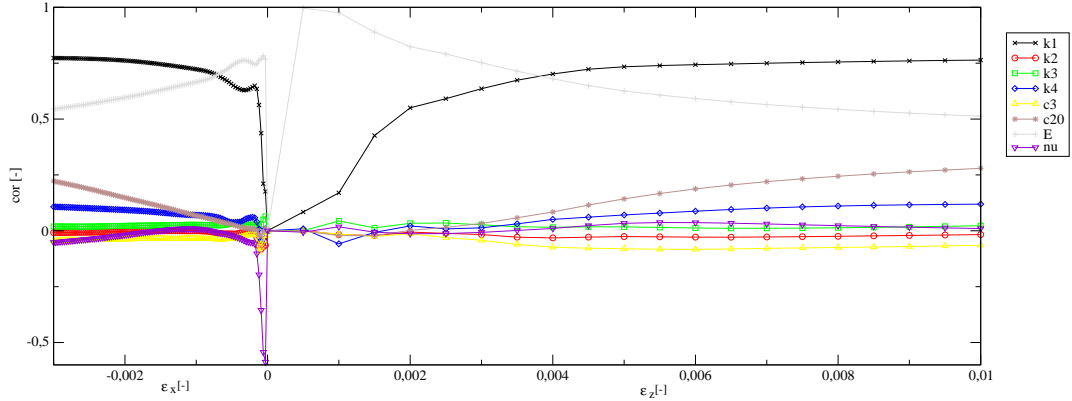}
\caption{Sensitivity evolution for uniaxial compression
test} 
\label{fig:uniaxial_compression_sensitivity}
\end{figure}

The LHS sampling procedure has been used to determine the set of 30
simulations resulting in a~``bundle'' of stress-strain curves shown in
Figure~\ref{fig:uniaxial_compression_lhs}. The evolution of stochastic
sensitivity during the loading process is depicted in
Figure~\ref{fig:uniaxial_compression_sensitivity}. The results
indicate that the most sensitive parameters are Young's modulus $E$,
the coefficient $k_1$, Poisson's ratio $\nu$ (especially for the
initial stages of loading) and, for the later stages of loading, the
coefficient $c_{20}$. Therefore, one can expect that only these
parameters can be reliably identified from this test.

Moreover, the impact of individual parameters on a position of a peak
of stress-strain curves is computed. The results of a~sensitivity
analysis using Pearson's product moment correlation coefficient of
peak coordinates [$\epsilon$,$\sigma$] are listed in
Table~\ref{tab-valec_sa-peaks}.  Results indicate particularly strong
influence of the $k_1$ parameter, which hopefully allows its reliable
determination.

\begin{table}[!ht]
\centering
\begin{tabular}{c|rr}
\hline
 & \multicolumn{2}{c}{\bf Pearson's coefficients}  \\
\bf Parameter & $\epsilon$ & $\sigma$\\
\hline 
$k_1$    & 0.968  & 0.709     \\
$k_2$    & 0.025  & 0.008  \\
$k_3$    & 0.015  & 0.030    \\
$k_4$    & 0.021  & 0.074   \\
$c_3$    & -0.019 & -0.020   \\
$c_{20}$ & 0.158  & 0.041   \\
$E$      & 0.004  & 0.684    \\
$\nu$    & 0.129  & 0.000  \\
\hline
\end{tabular}
\caption{Pearson's coefficient as a sensitivity measure of individual
parameters to the peak coordinates [$\epsilon$,$\sigma$] of stress-strain curves}
\label{tab-valec_sa-peaks}
\end{table}

Based on the results of sensitivity analysis, the neural network
training can be performed using a~nested strategy. First, {\em Young's
modulus} $E$ with sensitivity $\approx 1$ in the initial stage is
easily identified. To this end, a~three-layer ANN is used. In the
first layer only the neurons corresponding to the values of stresses
$\sigma_{z,1}$, $\sigma_{z,2}$ and $\sigma_{z,3}$ in the first three
points of axial strain with extremal Pearson's correlation coefficient
are chosen. The second layer contains two neurons only; the
last layer consists of one neuron corresponding to the predicted value
of Young's modulus $E$. For the ANN training the GRADE algorithm is
used and calculation is stopped after 1,000,000 iterations of the
algorithm.

The three-layer ANN trained using the {\bf GRADE} algorithm is also
used for the identification of other microplane model parameters. In
Table~\ref{tab_uni-net30} network's architectures and the choice of
input values for the identification of each microplane parameter are
presented.

\begin{table}[!ht]
\centering
\begin{tabular}{c|c|c}
\hline
\bf Parameter & \bf ANN's layout & \bf Input values\\
\hline 
$k_1$    & 4 - 2 - 1  & $\sigma_{x,2}$, $\sigma_{z,peak}$,
$\epsilon_{z,peak}$, prediction of $E$ \\ 
$k_2$    & 4 - 2 - 1  & $\sigma_{z,20}$, $\sigma_{z,peak}$,
$\epsilon_{z,peak}$, prediction of $E$ \\ 
$k_3$    & 4 - 2 - 1  & $\sigma_{z,20}$, $\sigma_{z,peak}$,
$\epsilon_{z,peak}$, prediction of $E$ \\ 
$k_4$    & 4 - 2 - 1  & $\sigma_{z,20}$, $\sigma_{z,peak}$,
$\epsilon_{z,peak}$, prediction of $E$ \\ 
$c_3$    & 4 - 2 - 1  & $\sigma_{z,20}$, $\sigma_{z,peak}$,
$\epsilon_{z,peak}$, prediction of $E$ \\ 
$c_{20}$ & 4 - 3 - 1  & $\sigma_{x,100}$, $\sigma_{z,20}$, prediction
of $E$, prediction of $k_1$   \\ 
$E$      & 3 - 2 - 1  & $\sigma_{z,1}$, $\sigma_{z,2}$, $\sigma_{z,3}$    \\
$\nu$    & 4 - 3 - 1  & $\sigma_{x,1}$, $\sigma_{x,2}$, prediction of $E$,
prediction of $k_1$ \\
\hline
\end{tabular}
\caption{Neural network architectures}
\label{tab_uni-net30}
\end{table}

The results of the identification for an independent set of ten
stress-strain curves using the proposed strategy are shown in
Table~\ref{tbl_err_valec30}.
\begin{table}[htb]
\centering
\begin{tabular}{c|ll|rr}
\hline
\bf Parameter & \multicolumn{2}{|c|}{\bf Absolute error} &
\multicolumn{2}{c}{\bf Relative error [\%]} \\ 
 & average & maximal & average & maximal \\
\hline
$k_1$    & 2.058e-06 & 4.678e-06 & 1.34 & {\bf 2.76}  \\    
$k_2$    & 138.9     & 318.7     & 38.92 & 179.62  \\ 
$k_3$    &  2.679    & 5.283     & 33.96 & 102.33  \\ 
$k_4$    & 52.70     & 91.70     & 48.33 & 107.17  \\ 
$c_3$    & 1.675     & 2.278     & 37.66 & 47.09  \\  
$c_{20}$ & 0.7547    & 1.4168    & 26.70 & 56.69  \\  
$E$      & 229.3     & 594.5     & 0.74 & {\bf 1.79} \\     
$\nu$    & 0.006447  & 0.010361  & 2.93 & {\bf 4.72}  \\    
\hline
\end{tabular}\
\caption{Errors in the estimated parameters obtained from ten
independent tests}
\label{tbl_err_valec30}
\end{table}

Note that obtained errors are in a close agreement with the results of
sensitivity analysis. Except $E$, $\nu$ and $k_1$, the parameters of
the model are identified with very high error values. Therefore,
additional simulations are needed to obtain these values reliably.

\begin{figure}[tbh]
\centering
\includegraphics*[width=8cm,keepaspectratio]{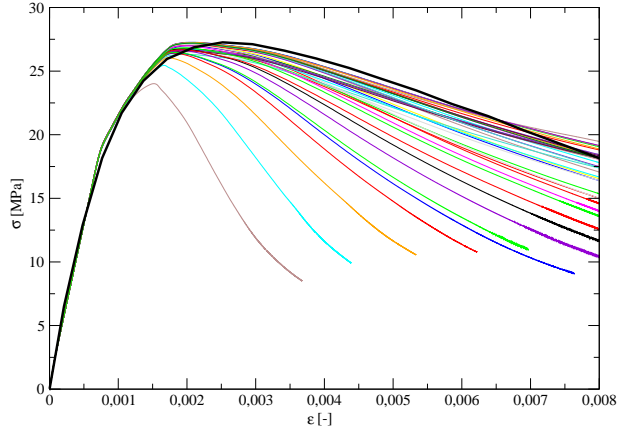}
\caption{Bundle of simulated stress-strain curves for uniaxial
compression with fixed values of Young's modulus, Poisson's ratio and
$k_1$ parameter and one (bold black) measured stress-strain curve}
\label{fig_uni-k1-bundle}
\end{figure}

\begin{figure}[tbh]
\centering
\includegraphics*[width=8cm,keepaspectratio]{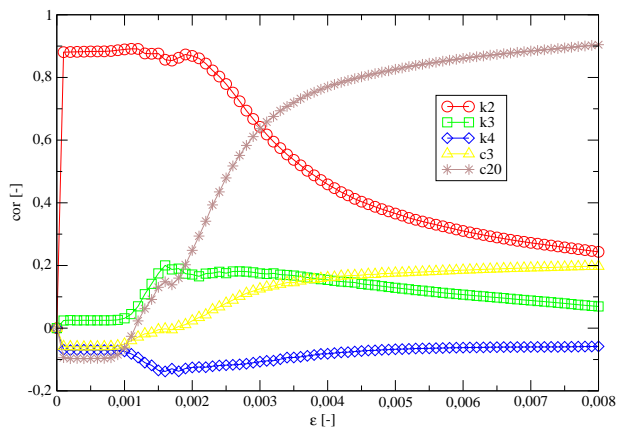}
\caption{Evolution of Pearson's correlation coefficient during the
loading test for fixed values of $E$, $\nu$ and $k_1$ parameters}
\label{fig_uni-k1-sens}
\end{figure}

At this point we have to fix already well-identified parameters to the
optimized values and perform simulations for the rest of
parameters. To minimize computational time, values for one uniaxial
measurement presented later in Section~\ref{sec:real} were
used. Corresponding values predicted by previously learned neural
networks were Young's modulus $E = 32035.5$~MPa and $k_1 =
0.000089046$. Poisson's ratio is set to $\nu = 0.2$ as a~usual value
of a wide range of concretes. Next, 40 new simulations varying the
rest of unknown parameters are computed. From this suite, only 34
solutions are valid, i.e. these solutions were able to reach the
desired value of axial strain $\strain = 0.008$. The bundle of
resulting curves is shown in Figure~\ref{fig_uni-k1-bundle}. Note that
the black bold curve represents measured data. The evolution of
Pearson's correlation coefficient during the experiment is shown in
Figure~\ref{fig_uni-k1-sens}.

\begin{figure}[tbh]
\centering
\includegraphics*[width=8cm,keepaspectratio]{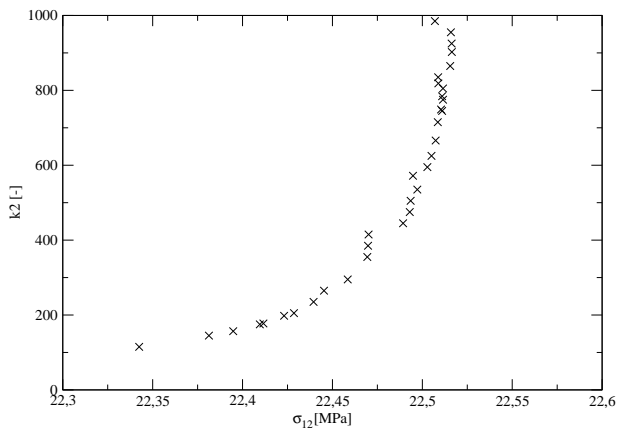}
\caption{$k_2$ parameter as a function of the stress $\sigma_{12}$
(corresponding to $\epsilon =0.0011$)}
\label{fig_k2-rez12}
\end{figure}

\begin{figure}[tbh]
\centering
\includegraphics*[width=8cm,keepaspectratio]{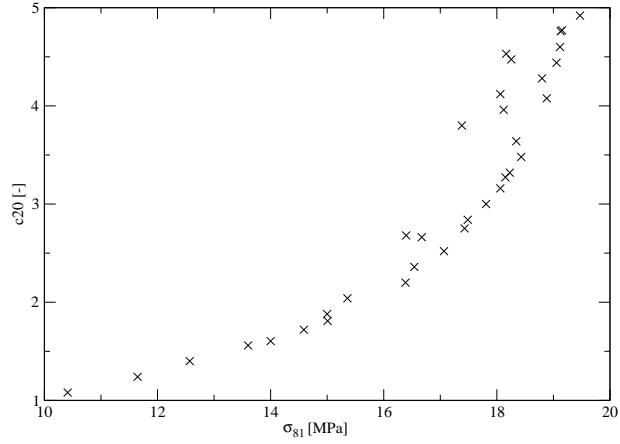}
\caption{The $c_{20}$ parameter as a function of a stress
($\sigma_{81}$) at the end of simulations}
\label{fig_c20-rez81}
\end{figure}

The sensitivity shows very high influence of $k_2$ parameter at the
beginning of the loading. If we inspect Figure~\ref{fig_k2-rez12}, it
is clear that the $k_2$ parameter influences the stress-strain curve
only on a very narrow interval and hence the parameter $k_2$ cannot be
identified from this test (the authors of the microplane model indeed
proposed $k_2$ to be estimated from the triaxial compression test).
We are more interested in fitting data in the post-peak part. For
post-peak curves, sensitivity analysis shows especially growing
influence of $c_{20}$ parameter. This can be demonstrated by a
relation between the $c_{20}$ parameter and a~value of a~stress
($\sigma_{81}$) at the end of our simulations, where the correlation
coefficient reaches the value 0.904429. This relation is graphically
illustrated in Figure~\ref{fig_c20-rez81}.

\begin{figure}[tbh]
\centering
\includegraphics*[width=80mm,keepaspectratio]{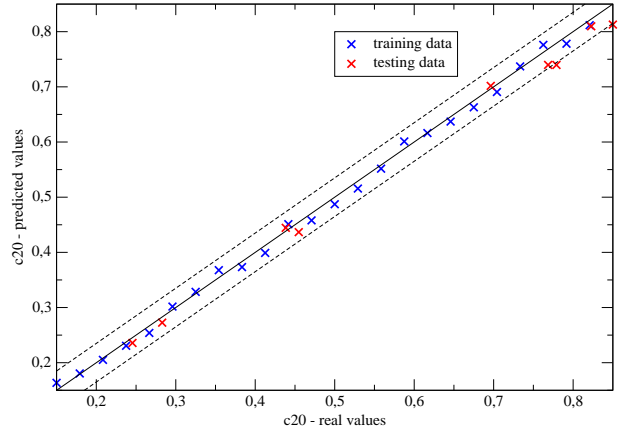}
\caption{Quality of ANN predictions of $c_{20}$ parameter}
\label{fig_c20-ann-pred}
\end{figure}

It is clearly visible that this relation is not highly non-linear and
any simple regression of this data is not possible. We applied the ANN
with 3 input neurons chosen to get the best prediction of parameters
based on post-peak curves. Therefore, one input value is a stress
value at the peak $\sigma_{peak}$ and the other two inputs are stress
values $\sigma_{61}$ and $\sigma_{81}$ corresponding to strains
$\epsilon = 0.006$ and $\epsilon = 0.008$, respectively. Two neurons
in the hidden layer were used. Quality of ANN prediction is
demonstrated in Figure~\ref{fig_c20-ann-pred}. In particular, the
exact prediction of the searched value corresponds to a point lying
on the the axis of the first quadrant. Values of predicted parameters
are normalized here to the interval $\langle 0.15, 0.85
\rangle$. Dashed parallel lines bound a~$5\%$ relative error related to
the size of the parameter's interval. Clearly, the identification
procedure works with an error which does not exceed the selected $5\%$
tolerance.

\begin{figure}[tbh]
\centering
\includegraphics*[width=80mm,keepaspectratio]{figures/errorsc20.eps} 
\caption{Evolution of ANN's errors during the training in prediction
of $c_{20}$ parameter}
\label{fig_c20-ann-err}
\end{figure}

The attention is also paid to the over-training of the ANN. To control
this aspect, the evolution of errors in ANN's predictions during the
training process on the training and testing data are monitored (see
Figure~\ref{fig_c20-ann-err}). Recall that if the errors on the
testing set are much higher than on the training set, we suppose such
an ANN to be over-trained. Even though this seems to be the case of
the current ANN, we attribute such a behavior to the fact that there
are more training data (in our case $25$) then $11$ neural weights
optimized by the algorithm. Also note the typical restarting of the
optimization process caused by the multi-modal optimization strategy
CERAF presented in Section~\ref{sec:GRADE}.

\subsection{Hydrostatic test}
\begin{figure}[tb]
\centering
\subfigure[]{
\begin{minipage}[b]{0.25\textwidth}
  \centering \includegraphics[height=5cm]{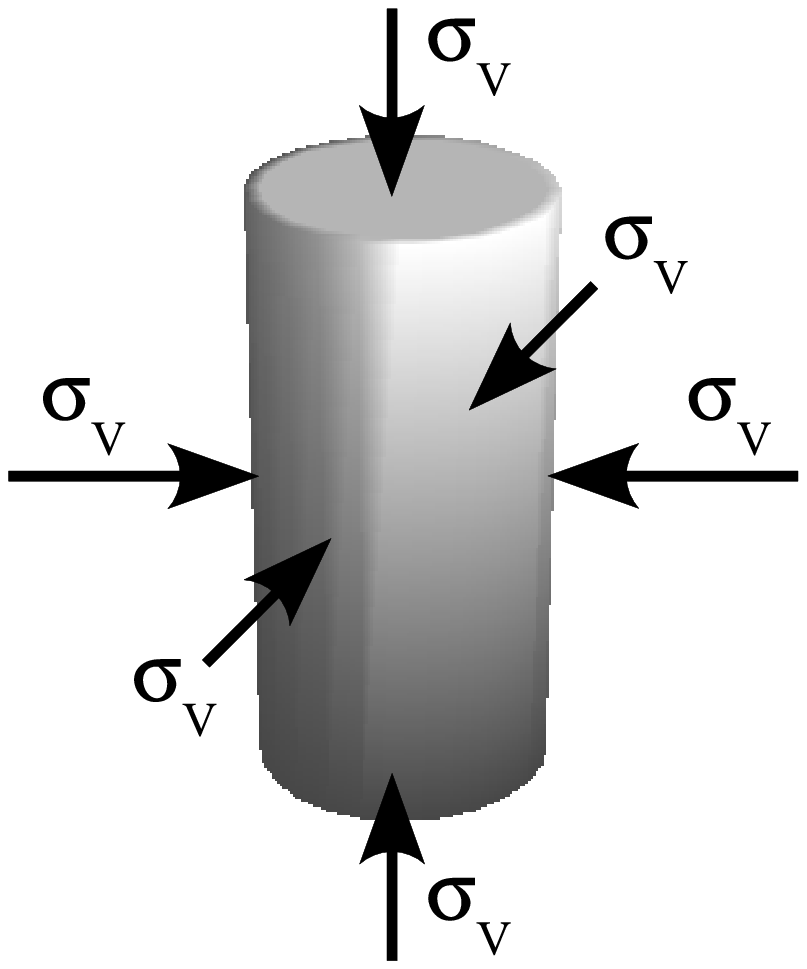}
\end{minipage}}%
\subfigure[]{
\begin{minipage}[b]{0.23\textwidth}
  \centering \includegraphics[height=4cm]{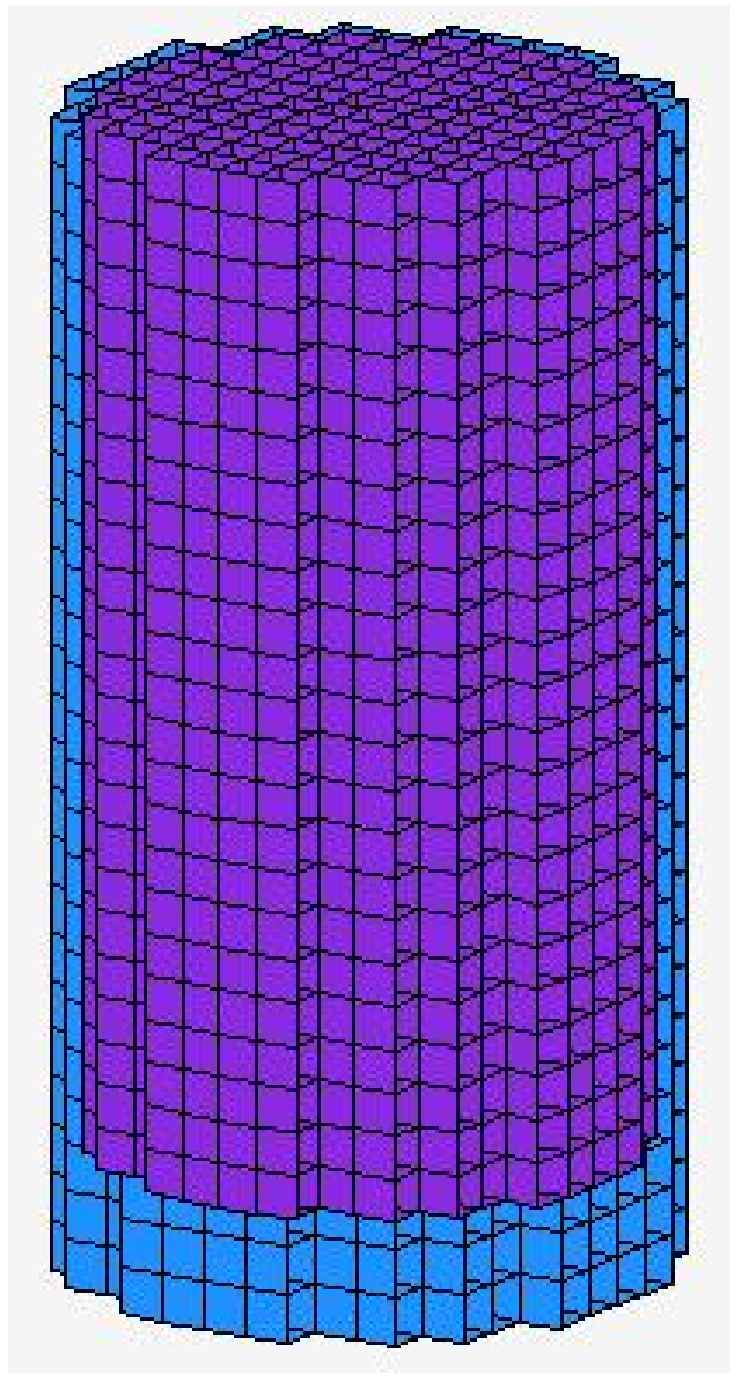}
\end{minipage}}%
\subfigure[]{
\begin{minipage}[b]{0.5\textwidth}
  \centering \includegraphics[height=5cm]{figures/hyd-bundle2.eps}
\end{minipage}}%
\caption{Hydrostatic test. (a)~Experiment setup, (b)~Initial and
deformed finite element mesh, (c)~Stress-strain curves}
\label{fig:hydrostatic_test}
\end{figure}

\begin{figure}[tbh]
\centering
\includegraphics*[width=12cm,keepaspectratio]{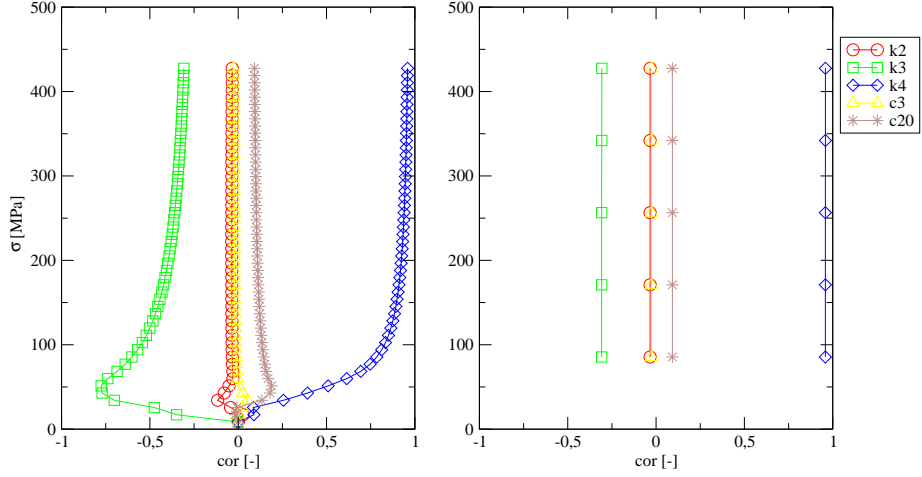}
\caption{Evolution of Pearson's correlation coefficient during the
hydrostatic compression test for loading (left) and unloading (right)
branch}
\label{fig:hydrostatic_test_sensitivity}
\end{figure}

%
The next independent test used for the identification problem is the
hydrostatic compression test, where a concrete cylinder is subjected
to an increasing uniform pressure. Axial and two lateral deformations
(in mutually perpendicular directions) are measured. The experimental
setup is shown in Figure~\ref{fig:hydrostatic_test}a--b.  Again, to
improve identification precision, the parameters $E$, $\nu$ and $k_1$
are supposed to be fixed to the previously identified values. The
``bundle'' of stress-strain curves obtained using the LHS sampling for
70 samples is depicted in Figure~\ref{fig:hydrostatic_test}c and the
corresponding sensitivity evolution in
Figure~\ref{fig:hydrostatic_test_sensitivity}. Note that the maximal
value of a hydrostatic pressure for all these tests is 427.5 MPa.

\begin{figure}[tbh]
\centering
\includegraphics*[width=8cm,keepaspectratio]{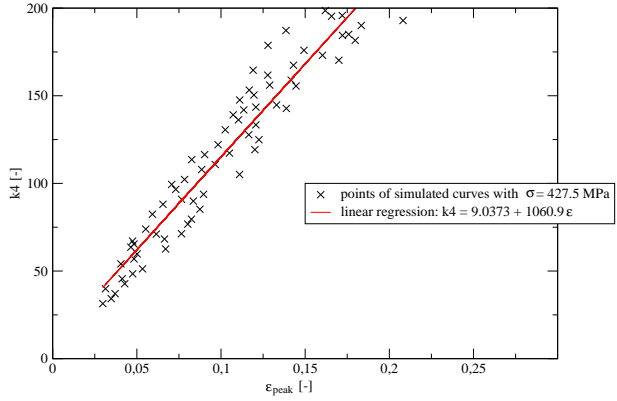}
\caption{$k_4$ parameter as a function of a strain of a peak}
\label{fig_k4-rez}
\end{figure}

The sensitivity information reveal that this test can be used to
identify parameter $k_3$ from the loading branch while a~combination
of loading and/or unloading data can be used for $k_4$ parameter
identification. Moreover, the correlation between the strain at the
peak of curves and $k_4$ parameter is so high that one can expect
their relation to be almost linear. This is, however, not the case as
illustrated by Figure~\ref{fig_k4-rez} showing the~value of $k_4$
parameter as a function of a~strain $\strain$. In spite of a high
value of the correlation coefficient equal to $0.958586$, the noise of
these data seems to be very high and a~use of a~linear regression
introduces a high error in $k_4$ parameter prediction.

\begin{figure}[tbh]
\centering
\includegraphics*[width=8cm,keepaspectratio]{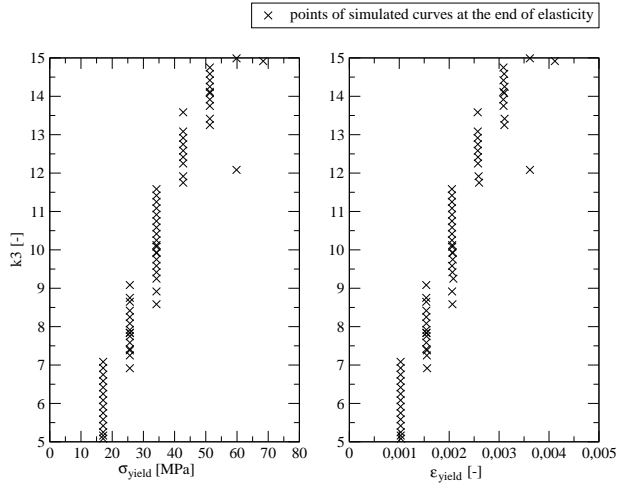}
\caption{$k_3$ parameter as a function of a position of the end of an
elastic stage}
\label{fig_k3-rez}
\end{figure}

To identify the $k_3$ parameter, which shows high correlation near to
the end of elastic stage, we evaluate the correlation coefficient
between this parameter and a position of the end of an elastic
stage. $k_3$ and $\epsilon_{yield}$ correlation is $0.95078$ and $k_3$
and $\sigma_{yield}$ = $0.951873$. In Figure~\ref{fig_k3-rez}, values
of $k_3$ parameter as a function of these coordinates are
presented. The discrete values of $\epsilon_{yield}$ are caused by the
size of a~time step at the beginning of simulations. The noise in data
is very high again and it is not possible to reliably use a~linear
regression. Because all other parameters have very small correlation,
we suppose that the noise of the parameter $k_3$ is caused by $k_4$
parameter and vice-versa. In other words, these noises could be caused
by some level of correlation between the parameters $k_3$ and
$k_4$. Hence we decided to apply an artificial layered neural network
again. The first 60 simulations prepared by the LHS method were used
for training and remaining (randomly chosen) 10 simulations for
testing. Particular choice of input values as well as architectures of
ANN's is shown in Table~\ref{tab_hyd-ann}. To eliminate unknown
correlation between parameters $k_3$ and $k_4$, their values are used
also as inputs into ANN's.

\begin{table}[!ht]
\centering
\begin{tabular}{c|c|c}
\hline
\bf Parameter & \bf ANN's layout & \bf Input values\\
\hline 
$k_3$    & 5 - 2 - 1  & $k_4$, $\epsilon_{yield}$, $\epsilon_{load,2}$, $\epsilon_{load,5}$,$\epsilon_{peak}$ \\
$k_4$    & 3 - 2 - 1  & $k_3$, $\epsilon_{peak}$, $\epsilon_{unload,4}$   \\
\hline
\end{tabular}
\caption{Neural network architectures for hydrostatic test}
\label{tab_hyd-ann}
\end{table}

\begin{figure}[tbh]
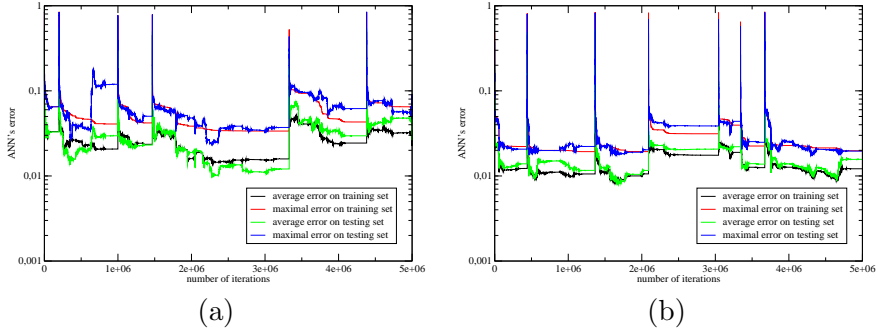

\centering
\begin{tabular}{cc}
\includegraphics*[width=55mm,keepaspectratio]{figures/errorsk3.eps} &
\includegraphics*[width=55mm,keepaspectratio]{figures/errorsk4.eps} \\
(a) & (b)
\end{tabular}
\caption{Evolution of ANN's errors during the training process in prediction
of (a) $k_3$ parameter and (b) $k_4$ parameter} 
\label{fig_ann-err}
\end{figure}

\begin{figure}[tbh]
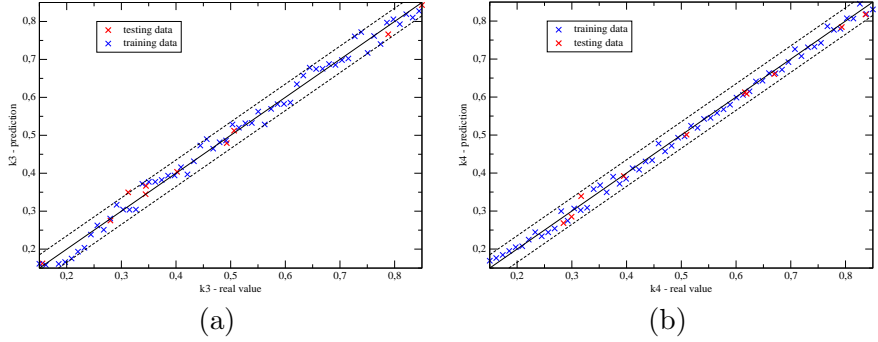

\centering
\begin{tabular}{cc}
\includegraphics*[width=55mm,keepaspectratio]{figures/k3-pred.eps} &
\includegraphics*[width=55mm,keepaspectratio]{figures/k4-pred.eps} \\
(a) & (b)
\end{tabular}
\caption{Quality of ANN prediction of (a) $k_3$ parameter and (b) $k_4$ parameter}
\label{fig_ann-pred}
\end{figure}

The architectures of ANN's were chosen manually to get the best
precision in predictions and also to avoid the over-training of the
ANN's. Therefore, it is possible to show the evolution of ANN's errors
(see Figure~\ref{fig_ann-err}) during the training. A training process
with 5000000 iterations takes approximately 20~minutes. Quality of
ANN's predictions is demonstrated in Figure~\ref{fig_ann-pred}. Values
of predicted parameters are again normalized into the interval
$\langle 0.15, 0.85 \rangle$.

\begin{figure}[tbh]
\centering
\includegraphics*[width=8cm,keepaspectratio]{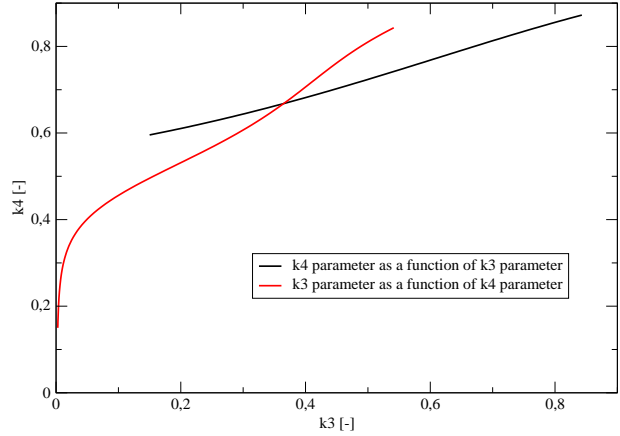}
\caption{Relations of $k_3$ and $k_4$ parameters}
\label{fig_k3k4}
\end{figure}

\begin{figure}[tbh]
\centering
\includegraphics*[width=8cm,keepaspectratio]{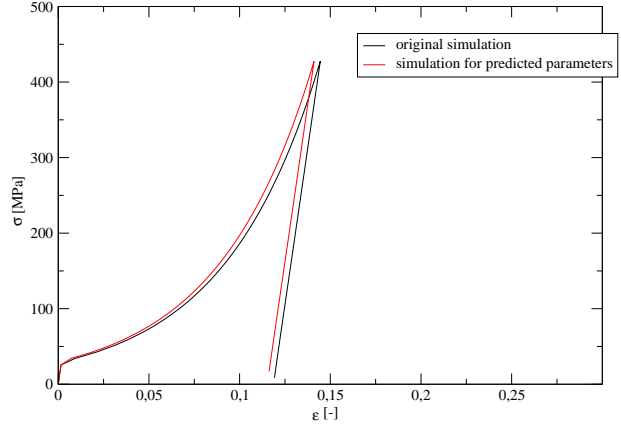}
\caption{Comparison of original simulation and simulation for
predicted $k_3$ and $k_4$ parameters}
\label{fig_70-compr}
\end{figure}

In this way, two ANN's or, in other words, two implicit functions are
prescribed. One defines a value of $k_3$ parameter depending on
a~value of $k_4$ parameter and some other properties of
a~stress-strain curve, the second define a~value of $k_4$ parameter
depending on a~value of $k_3$ parameter and some other properties of
a~stress-strain curve. Once we get some "measured" data and we fix all
properties of a~stress-strain curve, we get a~system of two non-linear
equations for $k_3$ and $k_4$. We can solve this system, e.g.,
graphically. Both relations are depicted in Figure~\ref{fig_k3k4} for
one independent stress-strain curve ($k_3 = 7.84293$, $k_4 =
155.551$). Their intersection defines searched parameters, $k_3 =
8.15687$ and $k_4 = 154.072$ in this particular case. The precision of
the proposed strategy is visible in comparison of corresponding
stress-strain curves, see Figure~\ref{fig_70-compr}. Note, that $E$,
$\nu$ and $k_1$ were the same as in previous section and remaining
parameters, i.e. $k_2$, $c_3$ and $c_{20}$, were chosen
randomly.\footnote{%
Theoretically, the value of $c_{20}$ is known in this stage
from the previous identification step. Nevertheless, the numerical
simulation of the unidirectional experiment takes substantially more
time to complete, see Section~\ref{sec:concl} for an example, and therefore the
independence of $c_{20}$ allows us to proceed with the
inverse analysis even though the first phase is not finished. 
}

\subsection{Triaxial test}
\begin{figure}[tb]
\centering
\subfigure[]{
\begin{minipage}[b]{0.32\textwidth}
  \centering \includegraphics[height=5cm]{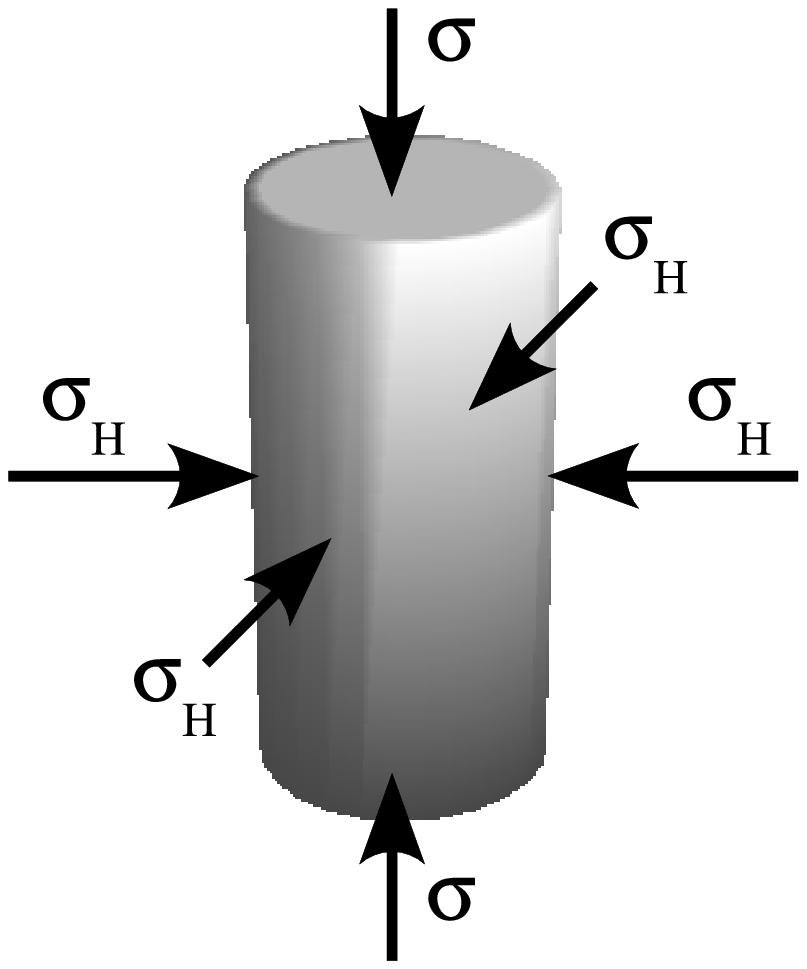}
\end{minipage}}%
\subfigure[]{
\begin{minipage}[b]{0.32\textwidth}
  \centering \includegraphics[height=4.5cm]{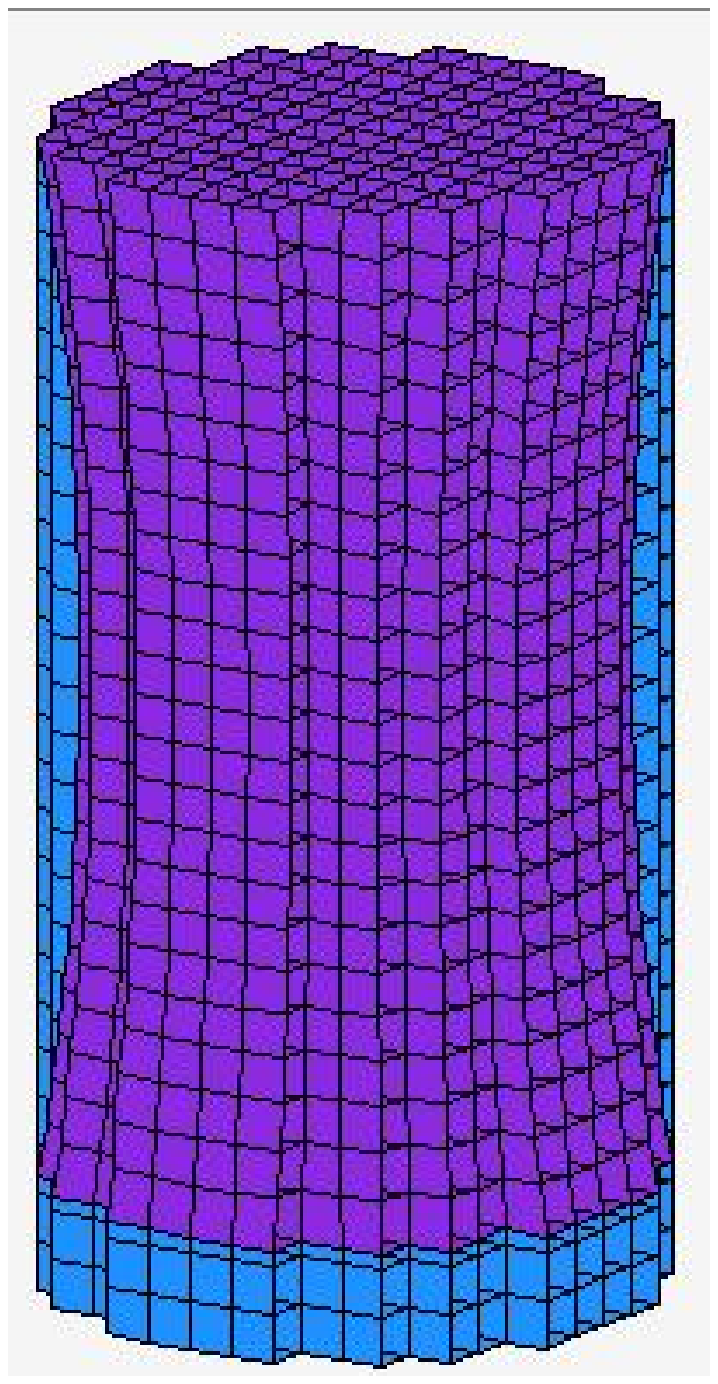}
\end{minipage}}%
\subfigure[]{
\begin{minipage}[b]{0.32\textwidth}
  \centering \includegraphics[height=4.5cm]{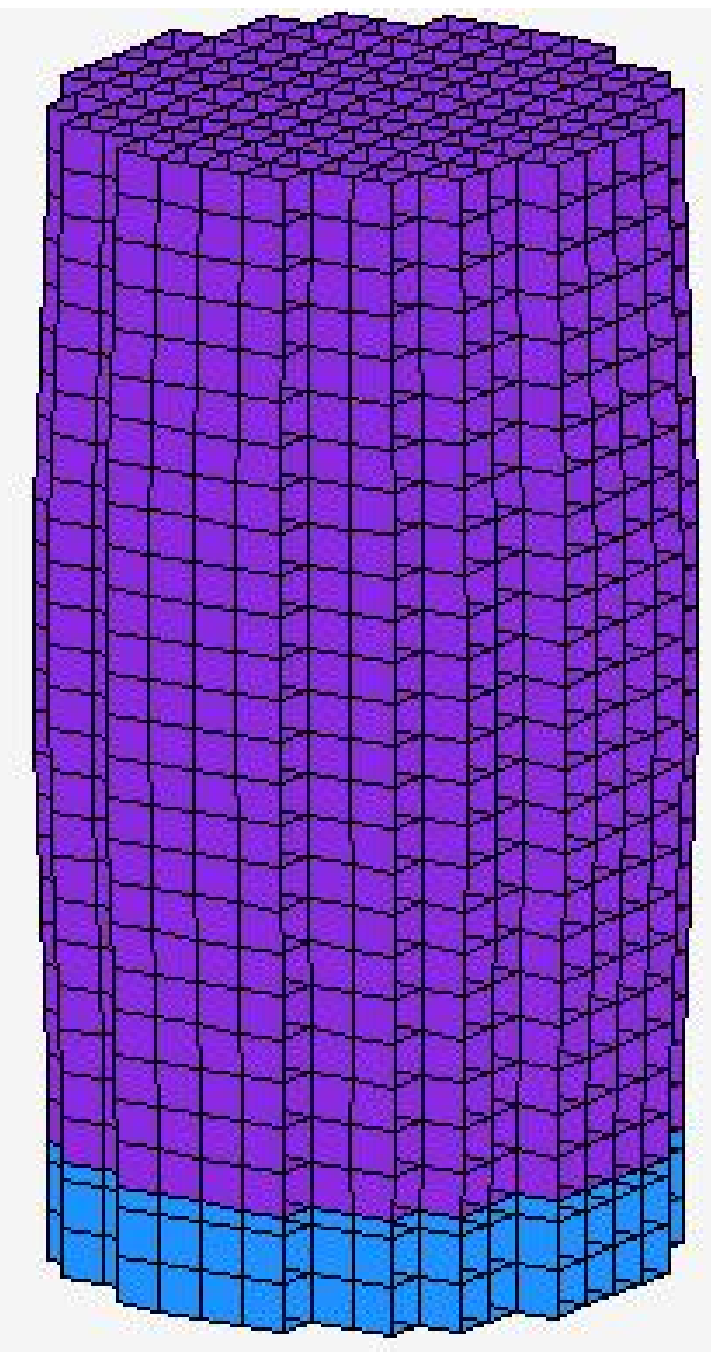}
\end{minipage}}%
\caption{Triaxial compression test. (a)~Experiment setup, (b)~Initial
and deformed mesh at the end of hydrostatic loading, (c)~Initial and
deformed mesh at the end of total loading}
\label{fig:triaxial_test}
\end{figure}

%
The last experiment, used for the purpose of parameter identification,
is a~triaxial compression test. To this end, a~specimen is subjected
to the hydrostatic pressure $\stress_H$. After the peak value of
$\stress_H$ is reached, the axial stress is proportionally
increased. The ``excess'' axial strain $\strain = \strain_T -
\strain_H$, where $\strain_T$ and $\strain_H$ denote the total and
hydrostatic axial strain, is measured as a~function of the
overall stress $\stress$. The test setup is shown in
Figure~\ref{fig:triaxial_test}.

\begin{figure}[tbh]
\centering
\includegraphics*[width=8cm,keepaspectratio]{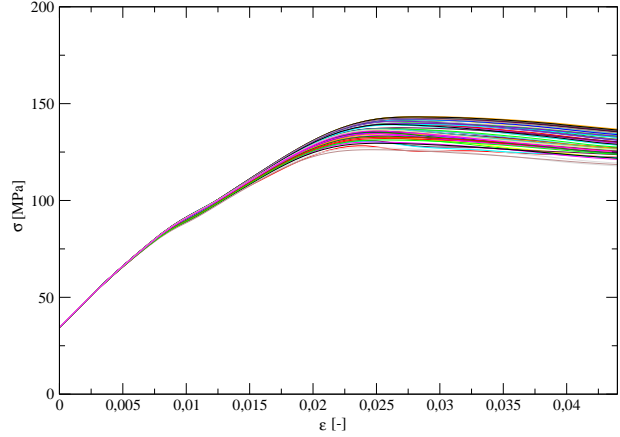}
\caption{Bundle of simulated stress-strain curves for triaxial compression test}
\label{fig_tri-bundle}
\end{figure}

\begin{figure}[tbh]
\centering
\includegraphics*[width=8cm,keepaspectratio]{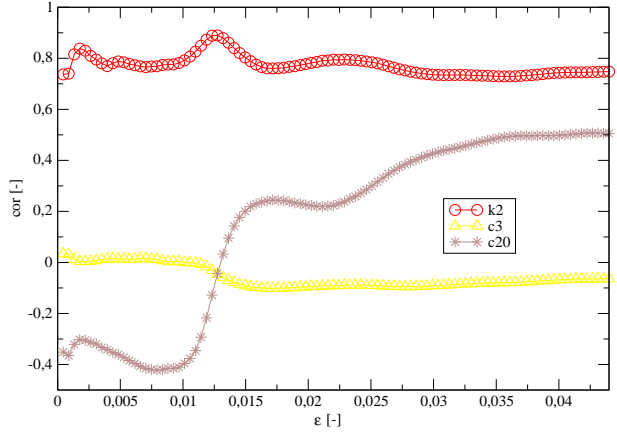}
\caption{Evolution of Pearson's correlation coefficient during the
triaxial compression test}
\label{fig_tri-sens}
\end{figure}

At this point, we assume that parameters $E, \nu$, $k_1$, $k_3$ and
$k_4$ are known from previous identifications\footnote{i.e. $E =
32035.5$~MPa, $\nu = 0.2$, $k_1 = 0.000089046$, $k_3 = 8.15687$ and
$k_4 = 154.072$.}. Next, 70 simulations (60 training and 10 testing)
of the triaxial compression test are computed by varying three
remaining parameters $k_2$, $c_3$ and $c_{20}$. The bundle of
stress-strain curves for $\stress_H = 34.5$~MPa is shown in
Figure~\ref{fig_tri-bundle} together with the evolution of Pearson's
correlation coefficient during the experiment in
Figure~\ref{fig_tri-sens}.


\begin{table}[!ht]
\centering
\begin{tabular}{c|rr}
\hline
 & \multicolumn{2}{c}{\bf Pearson's coefficient}  \\
\bf Parameter & $\epsilon$ & $\sigma$\\
\hline 
$k_2$    & 0.585  & 0.791  \\
$c_3$    & -0.067  & -0.088   \\
$c_{20}$ & 0.664  & 0.329   \\
\hline
\end{tabular}
\caption{Pearson's coefficient as a~sensitivity measure of individual
parameters to the peak coordinates [$\epsilon$,$\sigma$] of
stress-strain curves}
\label{tab_tri-sens-peaks}
\end{table}

\begin{figure}[tbh]
\centering
\includegraphics*[width=8cm,keepaspectratio]{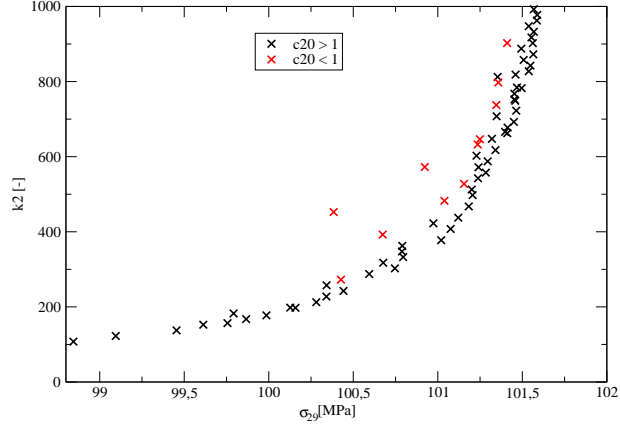}
\caption{$k_2$ parameter as a function of the stress value $\sigma_{29}$}
\label{fig_k2-rez}
\end{figure}

In addition, the correlation coefficient between microplane parameters
and stress and strain values of peaks is computed. These results are
shown in Table~\ref{tab_tri-sens-peaks}. It is visible that maximal
correlation is between $k_2$ parameter and the value of the stress
$\sigma_{29}$ corresponding to the strain equal to $\epsilon_{29} =
0.01276$. This correlation is 0.88956 and at the same time, the
correlation between these $\sigma_{29}$ values and $c_{20}$ parameter
is very small, therefore $c_{20}$ parameter does not influence the
relation between $k_2$ parameter and
$\sigma_{29}$. Figure~\ref{fig_k2-rez} shows that only small values of
$c_{20}$ parameter disturb this relation. In particular, points
related to the $c_{20}$ parameter smaller then $1$ lead to
oscillatory dependence.

\begin{figure}[tbh]
\centering
\includegraphics*[width=80mm,keepaspectratio]{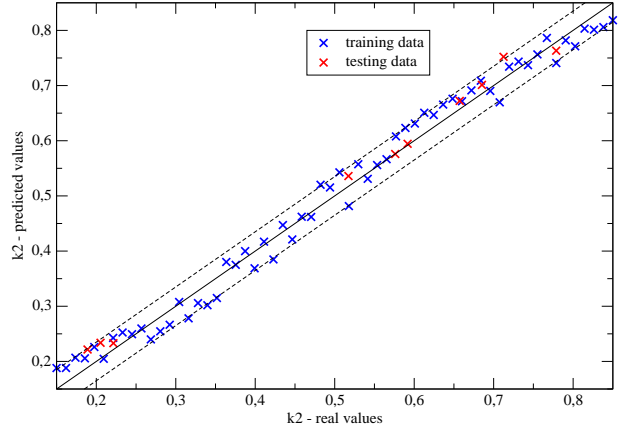} 
\caption{Quality of ANN prediction of $k_2$ parameter.}
\label{fig_tri-ann-pred}
\end{figure}

\begin{figure}[tbh]
\centering
\includegraphics*[width=80mm,keepaspectratio]{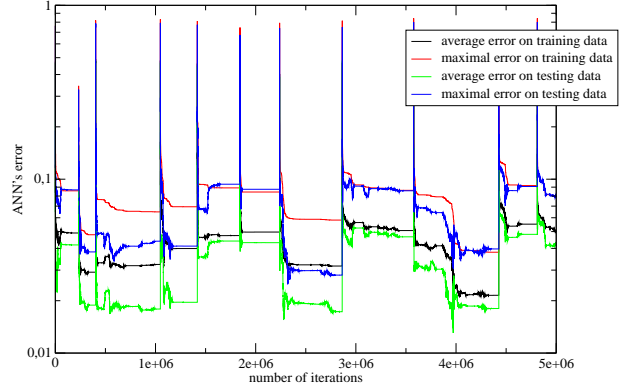} \\
\caption{Evolution of ANN's errors during the training in prediction
of $k_2$ parameter}
\label{fig_tri-ann-err}
\end{figure}

Because the highest correlation for $k_2$ parameter is again at the
beginning of the loading and after our experiences in identification
parameters from the uniaxial compression test, we were again afraid of
small significance of $k_2$ parameter to the shape of curves. Also
$\sigma_{29}$ does not seem to be significant. Therefore we made
several short computations with randomly chosen fixed value of
$c_{20}$ parameter to filter out its influence. We have got a~bundle
of curves showing similar spread of values as curves in
Figure~\ref{fig_tri-bundle}.  Therefore, it can be concluded that
these differences are probably caused by $k_2$ parameter only and a
neural network for $k_2$ parameter identification can be
designed. Because the bundle of curves varies mostly in the post-peak
part and we would like to get a predictor capable to fit this part of
a~curve properly, we use $\sigma_{peak}$ and $\sigma_{100}$ as input
values. The latter one correspond to the end of our simulations, where
$\epsilon = 0.044$. We also add the third input value -- $\sigma_{29}$
-- because of its small correlation with $c_{20}$ parameter. Two
neurons in the hidden layer were used. Quality of the ANN prediction
is demonstrated in Figure~\ref{fig_tri-ann-pred}. Under- and
over-fitting issues were again checked by errors evaluations during
the training process, see Figure~\ref{fig_tri-ann-err}.


\begin{figure}[tbh]
\centering
\includegraphics*[width=8cm,keepaspectratio]{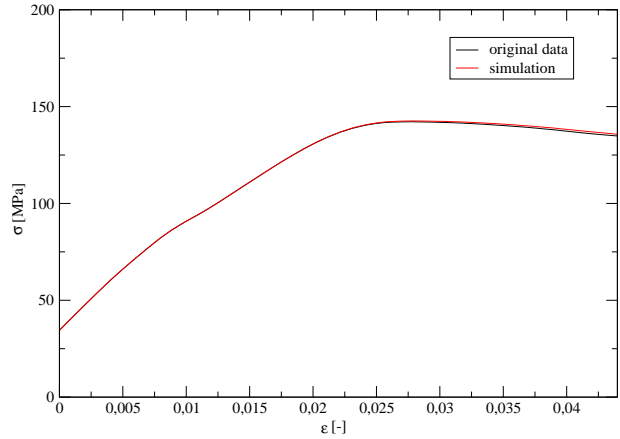}
\caption{Comparison of original simulation and simulation for
predicted parameters of triaxial compression test}
\label{fig_tri-70-compr}
\end{figure}

Almost perfect precision of the proposed strategy is visible in
comparison of corresponding stress-strain curves for $k_2 = 748.857$
and its prediction equal to $767.777$ and randomly chosen parameters
$c_{20}$ and $c_3$, see Figure~\ref{fig_tri-70-compr}.

\subsection{Application to measured data}\label{sec:real}
\begin{figure}[tbh]
\centering
\includegraphics*[width=8cm,keepaspectratio]{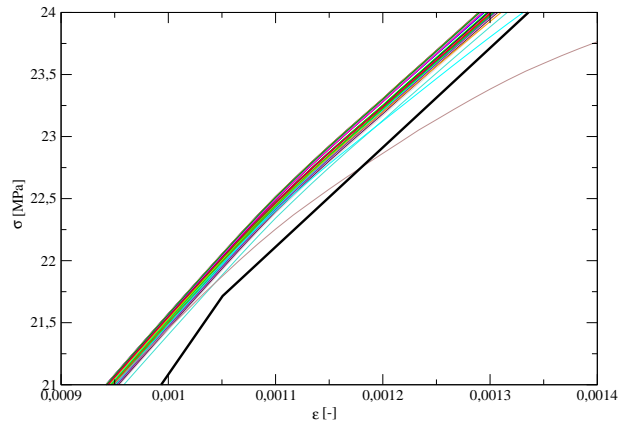}
\caption{Bundle of simulated stress-strain curves for uniaxial
compression and one (bold black) measured stress-strain curve under
zoom}
\label{fig_uni-k1-bundle-zoom}
\end{figure}

\begin{figure}[tbh]
\centering
\includegraphics*[width=8cm,keepaspectratio]{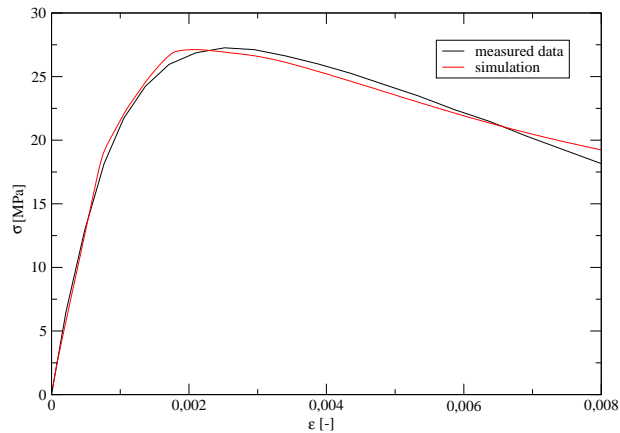}
\caption{Comparison of measured data and results of final simulation.}
\label{fig_uni+k1-compr}
\end{figure}

%
In previous sections, we have shown that the proposed methodology is
able to identify all but one ($c_3$) parameters from
computer-simulated curves. To demonstrate the applicability of the
proposed procedure, a~real simulation should be examined. However,
only limited experimental data from uniaxial compression tests are
available to authors which leaves us with only one uniaxial
stress-strain curve to be identified. As was mentioned previously in
Section~\ref{sec:uni}, Young's modulus $E = 32035.5~$~MPa, Poisson's
ratio $\nu = 0.2$ and $k_1 = 0.000089046$ are predicted by the neural
network for this measurement. Next, 30 samples for parameter $c_{20}$
are computed, see Figure~\ref{fig_uni-k1-bundle}. If we zoom into the
loading part of a stress-strain curve, it is clear, that the real
measurement is under all simulated data, see
Figure~\ref{fig_uni-k1-bundle-zoom}.  This part is influenced by high
correlation of $k_2$ parameter and therefore, it is clear that the
$k_2$ parameter cannot be obtained from this test. Finally we applied
our trained ANN to predict the $c_{20}$ parameter for the measured
data and we have got $c_{20} = 5.27065$. This value is out of the
interval specified for this parameter, but it is not surprising since
it is visible in Figure~\ref{fig_uni-k1-bundle} that measured data
somewhat deviate from the simulated bundle of curves. The final
comparison of measured data and a~simulation for predicted values of
$E$, $\nu$, $k_1$ and $c_{20}$ parameters is shown in
Figure~\ref{fig_uni+k1-compr}. The rest of unknown parameters are same
as in previous sections.


\section{Conclusion and future work}\label{sec:concl}
In the present contribution, an example of the engineering problem,
which is difficult to be solved by traditional procedures, was
solved using methods of soft computing. Particularly,
an artificial neural network was used to estimate required microplane
material model parameters. As the training procedure, the genetic
algorithm-based method {\bf GRADE} was used. A~number of needed
simulations is reduced by the application of the Latin Hypercube
Sampling method. The sensitivity analysis shows not only the influence
of individual parameters but also approximately predicts the errors
produced by the neural network.
\begin{table}[ht]
\centerline{
\begin{tabular}{c|c|c}
\hline
\bf Parameter & \bf Test & \bf ANN's topology \\
\hline
$E$           & Uniaxial compression & $3 + 2 + 1$\\
$\nu$         & Uniaxial compression & $4 + 3 + 1$\\
$k_1$         & Uniaxial compression & $4 + 2 + 1$\\
$k_2$         & Triaxial loading     & $3 + 2 + 1$ \\
$k_3$         & Hydrostatic loading  & $5 + 3 + 1$ \\
$k_4$         & Hydrostatic loading  & $3 + 2 + 1$ \\
$c_3$         & $\times$ & $\times$ \\
$c_{20}$      & Uniaxial compression & $3 + 2 + 1$\\
\hline 
\end{tabular}
}
\caption{Final status of M4 identification project}
\label{t:overview}
\end{table}

Results of our contribution, see Table~\ref{t:overview}, confirm the
claims made by authors~\cite{Bazant:2000:M4} of the microplane~M4
model on individual parameters fitting. Only the parameter $c_3$
remains undetermined but the parameter $c_3$ should be almost constant
for the wide range of concretes and our computations confirm almost zero
impact of this parameter on stress-strain curves.

The rather severe disadvantage of the microplane model, and also of
the proposed methodology, is an extreme demand of computational
time. A~suite of 30 uniaxial tests consumes approximately 25 days on
a~single processor PC with the Pentium~IV 3400~MHz processor and 3~GB
RAM. If we run tests in parallel on 7 computers, the needed time is
less than 4 days. The hydrostatic and triaxial tests are less
demanding, by running in parallel on 7 computers the required time is
less than one day for each test. Although several neural networks were
created to avoid the time-consuming numerical analysis, the proposed
methodology still needs to compute 30 uniaxial tests to properly
identify $c_{20}$ parameter and a set of 30 hydrostatic and triaxial
tests to fit $k_3$, $k_4$ and $k_2$. This drawback will be the subject
of future work.

\section*{Acknowledgments}
We would like to express our thanks to David Lehk\'{y} and
Drahom\'{\i}r Nov\'{a}k, Technical University of Brno, for numerous
suggestions and inspiring discussions. This work was supported by CEZ
MSM 6840770003 project.


\end{document}